\pdfoutput=1

\documentclass[11pt]{article}
\usepackage{adjustbox}
\usepackage{acl2023}

\usepackage{times}
\usepackage{latexsym}

\usepackage[T1]{fontenc}

\usepackage{microtype}

\usepackage{graphicx}
\usepackage{booktabs}
\usepackage{amsmath}
\usepackage{multirow}
\usepackage{float}
\usepackage{xspace}
\usepackage{caption}
\usepackage{subcaption}
\usepackage{array}
\usepackage{arydshln}
\usepackage{inconsolata}
\usepackage{titlesec}
\usepackage{siunitx}
\usepackage{hyperref}

\newcommand{\tp}[0]{\textsf{TP}\xspace}
\newcommand{\ghg}[0]{\textsf{GHG}\xspace}
\newcommand{\h}{\ensuremath{\mathcal{H}}\xspace}

\title{From Outliers to Topics in Language Models:\\ Anticipating Trends in News Corpora}

\author{\bf Evangelia Zve, {\bf Benjamin Icard,} {\bf Alice Breton,} \\ 
{\bf Lila Sainero,} {\bf Gauvain Bourgne,} {\bf and Jean-Gabriel Ganascia} \vspace{0.1in}\\ 
LIP6, Sorbonne University, CNRS, France}

\begin{document}
\maketitle

\begin{abstract}
This paper examines how outliers, often dismissed as noise in topic modeling, can act as weak signals of emerging topics in dynamic news corpora. Using vector embeddings from state-of-the-art language models and a cumulative clustering approach, we track their evolution over time in French and English news datasets focused on corporate social responsibility and climate change. The results reveal a consistent pattern: outliers tend to evolve into coherent topics over time across both models and languages.
\end{abstract}

\section{Introduction}
\label{sec:introduction}

As information ecosystems become increasingly dynamic, the early identification of emerging trends in news media remains a key challenge for natural language processing. Topic modeling, which clusters semantically similar documents to uncover latent themes, plays a central role in this task. Early approaches, most notably Latent Dirichlet Allocation (\texttt{LDA})~\cite{blei2003latent}, introduced a probabilistic framework to infer latent topics from textual documents~\cite{hoyle2022coherence}. More recent embedding-based methods, such as \texttt{BERTopic}~\cite{grootendorst2022}, represent documents as dense vector embeddings, enabling more contextualized representations and yielding more coherent topics in dynamic corpora such as online news content~\cite{babalola2024comprehensive}.

Unlike partition-based clustering methods often used for clustering vector embeddings, such as \texttt{KMeans}~\cite{hartigan1979kmeans}, or probabilistic topic models like \texttt{LDA}, both of which assign every document to a topic, \texttt{HDBSCAN}~\cite{campello2015hierarchical} is a density-based clustering algorithm that explicitly labels low-density points as \textit{outliers}. These documents, which do not fit into any topical cluster, are often treated as noise and excluded from downstream analysis.

Challenging the assumption that outliers are mere noise, we explore the hypothesis that outliers, documents not assigned to any cluster, may serve as early signals of emerging topics. We employ a cumulative clustering approach using \texttt{BERTopic} with \texttt{HDBSCAN}, tracing how isolated documents evolve and whether they are gradually integrated into clusters as their narratives gain salience. To aid interpretability, we also analyze lexical and stylistic features of outliers and their role in cluster integration.

To conduct our analysis, we use two news corpora. The first, in French, is a manually curated dataset documenting a corporate social responsibility dispute which serves as a pilot study. The second, in English, focuses on climate change and is used for replication. Both corpora are topically constrained, span continuous time periods, and provide full-text coverage, allowing to control for topical and timeline gaps.%

Section~\ref{sec:related} reviews related work. Section~\ref{sec:generalsetting} details the full experimental setting, with a particular focus on the methodology. Section~\ref{sec:pilot} presents the French study and results on outlier conversion. Section~\ref{sec:replication} reports replication results in English. Findings in both languages are discussed and compared in Section~\ref{sec:discussion}. Section~\ref{sec:conclusion} concludes and outlines future directions.

\section{Related Work}
\label{sec:related}

Topic modeling is widely applied across various domains, including corporate social responsibility~\cite{lee2023esg} and climate change~\cite{yla2022topic}, in both traditional and social media contexts~\cite{laureate2023systematic}. The field's methodological evolution, from probabilistic approaches like \texttt{LDA}~\cite{blei2003latent} to embedding-based methods such as \texttt{BERTopic}~\cite{grootendorst2022}, has improved semantic coherence. However, while outliers have been often treated as noise~\cite{alattar2021emerging}, their role in signaling emerging topics remains an underexplored area of research.

Research in temporal topic analysis has evolved from early techniques like burst detection~\cite{kleinberg2003bursty} and term-frequency-based change point identification~\cite{cataldi2010emerging} to more recent approaches tracking semantic drift~\cite{christophe2021monitoring} and transformer-based dynamic modeling~\cite{karakkaparambil2024evaluating,boutaleb2024bertrend}. While these methods effectively capture shifts in established topics, they typically overlook sparse outliers, documents that may precede and predict emerging themes before they coalesce into detectable clusters.

This relates to clustering methodology. While probabilistic topic models like \texttt{LDA} assign soft cluster memberships, and partition-based algorithms such as \texttt{KMeans}~\cite{hartigan1979kmeans} enforce hard assignments, both approaches assume that every document belongs to a cluster. In contrast, density-based methods like \texttt{HDBSCAN}~\cite{campello2015hierarchical} and \texttt{OPTICS}~\cite{ankerst1999OPTICS} explicitly identify outliers as low-density points that do not belong to any cluster. Unlike general anomaly detection techniques (e.g., Isolation Forest~\cite{liu2008isolation}, Local Outlier Factor~\cite{breunig2000lof}), which detect outliers without considering the topical coherence of thematically structured corpora, \texttt{HDBSCAN}’s built-in outlier detection aligns more closely with semantic structure. This allows  to track how semantically isolated documents may evolve into coherent topic clusters over time.

This paper examines whether outliers can serve as early signals of emerging topics. By tracking their integration into clusters over time via cumulative clustering, we aim to complement existing work focused on stable topic structures.

\section{Experimental Setting}
\label{sec:generalsetting}

\subsection{Hypothesis}
\label{ssec:mainhypotheses}

While topic modeling and document clustering have been extensively studied, the role of outliers in the dynamic formation of topics has not yet been explored. To address this gap, we propose the following hypothesis:

\begin{itemize}
    \item[\h:] \textit{In topic-based cumulative clustering of news articles, topics emerge or are reinforced in part through the assimilation of outliers—that is, documents initially unclustered that later become part of coherent topic clusters.}
\end{itemize}

This hypothesis assumes that topic formation in cumulative clustering reflects a gradual process of semantic integration, in which outliers may act as early signals of emerging or evolving topics.

\subsection{Models}
\label{ssec:models}

To test \h, we use nine open-source embedding models with diverse transformer architectures and language capabilities. Model selection was guided by performance on the Massive Text Embedding Benchmark (MTEB) \cite{muennighoff2022mteb}, as reported on the Hugging Face leaderboard\footnote{\url{https://huggingface.co/spaces/mteb/leaderboard}} as of September 16, 2024. Table~\ref{tab:detailmodels} (Appendix~\ref{sec:modelappendix}) summarizes the selected models.

\subsection{Methodology}
\label{ssec:methodology}

The methodology involves four main steps. First, we project news articles into a semantic space using language model embeddings. We then apply dimensionality reduction to enable efficient clustering and address the \textit{curse of dimensionality}~\cite{koppen2000curse}. Subsequently, we perform cumulative clustering over 20 monthly time windows and evaluate clustering quality to determine the optimal experimental configuration. Based on this setup, we test \h concerning outlier-to-topic conversion and assess its robustness through inter-model agreement. Finally, we analyze lexical and stylistic features to interpret differences between converted and non-converted outliers.

\subsubsection{Data Preparation}
\label{sssec:representation}

Each news article is represented using dense vector embeddings generated from nine pre-trained language models. For each document, we compute embeddings from three variants: body text, headline, and full article (both headline and body text). This projects articles into a high-dimensional semantic space, where distances reflect semantic similarity. We apply Uniform Manifold Approximation and Projection (\texttt{UMAP})~\cite{mcinnes2018umap} to reduce the dimensionality of embeddings prior to clustering. Output dimensions are varied across 2D, 3D, 5D, and 10D. \texttt{UMAP} is chosen over Principal Component Analysis (\texttt{PCA})~\cite{wold1987principal} due to its ability to preserve both local and global structure, which is important for identifying fine-grained topic distinctions and local outliers~\cite{TVCG.2023.3326569}.

\subsubsection{Cumulative Clustering}
\label{sssec:clustering}

We employ cumulative clustering (iterative topic modeling over expanding time windows) across 20 monthly intervals. At each step, documents from the current and all prior months are clustered jointly using \texttt{BERTopic} with \texttt{HDBSCAN}~\cite{mcinnes2017HDBSCAN}. This density-based algorithm assigns documents to clusters or labels them as outliers via the \texttt{GLOSH} algorithm~\cite{campello2015hierarchical}, which identifies low-density regions by comparing a point’s local density to its neighbors. Documents labeled \texttt{-1} are classified as outliers and excluded from clusters. To test \h, we track whether these outliers transition to inliers (i.e., join a cluster) in subsequent windows, thereby signaling emergent topics.

The clustering quality is evaluated using the silhouette score~\cite{shahapure2020cluster}, which measures cluster cohesion and separation. Scores above 0.7 are considered strong, 0.5–0.7 moderate, and below 0.25 weak. To evaluate clustering over time, we compute the mean and median silhouette scores across all time windows, and then aggregate these globally across all models. We compare the nine selected embedding models, content variants (headline, body, full article), and \texttt{UMAP} settings to ensure robustness. Based on these comparisons, we select the configuration with the highest silhouette score and proceed with our methodology to verify our hypothesis.

\vspace{-0.50em}
\subsubsection{Outlier-to-Topic Conversion}
\label{ssec:outtopic}

Under hypothesis \h, we evaluate whether outliers contribute to the formation of new topics or the reinforcement of existing ones. We compute, for each model, the proportion of outliers that later become topic inliers, and assess robustness via the rescaling method of \citet{icard2024multi}, which measures whether \h is consistently validated for the \textit{same} outliers across models. Specifically, for articles identified as outliers by \textit{all} models (at some point in their time window), we compute the proportion \( x \) of models that validate \h and rescale it as \( a = |2x - 1| \). This transformation captures consensus independently of polarity (as in Cohen’s kappa): both \( x = 1 \) (unanimous validation) and \( x = 0 \) (unanimous rejection) yield maximal agreement \( a = 1 \), while \( x = 0.5 \) corresponds to minimal agreement \( a = 0 \), since models are evenly split in this case.

\subsubsection{Lexicon and Writing Style Analysis}
\label{ssec:stylemethod}

As an attempt to explain the conversions observed, we first controlled for potential topical differences between converted and non-converted outliers using word-level \texttt{TfidfVectorizer} scores~\cite{qaiser2018text}, hereafter referred to as TF-IDF. Let \( w \) be a word and let \( \text{TFIDF}_g(w) \) denote its average TF-IDF score in group \( g \in \{\mathcal{H}, \text{not } \mathcal{H}\} \), where \(\mathcal{H}\) corresponds to outliers that were integrated into topic clusters (``converted''), and \(\text{not } \mathcal{H}\) to those that remained isolated (``non-converted''). To capture the differential lexical salience of word \( w \) across the two groups, we define the delta TF-IDF as:

\vspace{-1em}
\begin{equation}
\small
\Delta \text{TFIDF}(w) = \text{TFIDF}_{\mathcal{H}}(w) - \text{TFIDF}_{\text{not } \mathcal{H}}(w)
\label{eq:delta_tfidf}
\end{equation}
\vspace{-1em}

In addition, we investigated variation beyond lexical content by analyzing stylistic differences between converted and non-converted outliers using the stylometric framework introduced by~\citet{terreau2021writing}, which quantifies eight core stylistic dimensions. These include the relative frequency of \textit{function words} (e.g., prepositions, conjunctions, auxiliaries), \textit{punctuation marks} (e.g., periods, commas), \textit{numbers}, and \textit{named entities} (e.g., persons, organizations) per sentence; distributions of \textit{part-of-speech tags} (e.g., nouns, verbs, adjectives); and averages of \textit{structural features} (e.g., word length, word frequency, syllables per word). The framework also incorporates \textit{lexical complexity metrics} (e.g., Yule’s K~\cite{yule2014statistical}, Shannon entropy~\cite{shannon1948mathematical}) and \textit{readability indices} (e.g., Flesch-Kincaid Grade Level~\cite{kincaid1975derivation}).

\section{Pilot Study}
\label{sec:pilot}

\subsection{French Dataset} 

We constructed a dataset for the pilot study, referred to as \tp, consisting of 102 French news articles that we manually collected and curated. The articles document a controversy involving the major energy company \textit{TotalEnergies} and the prestigious French Grande École \textit{École Polytechnique}, who planned to build a research center on the university’s Saclay campus. The project drew both support, citing its contribution to energy research, and criticism, focused on academic independence and environmental impact. The \tp dataset covers the full timeline of media coverage, from December 2018 to August 2024, and includes documents from official sources, mainstream media, partisan outlets, opinion sections, and NGOs. It captures the entire development of the story, without topical or temporal gaps.%

\subsection{Topic-Based Clustering}
\label{ssec:tbc-tp}

\begin{figure}[h!]
    \centering
    \includegraphics[width=1\linewidth]{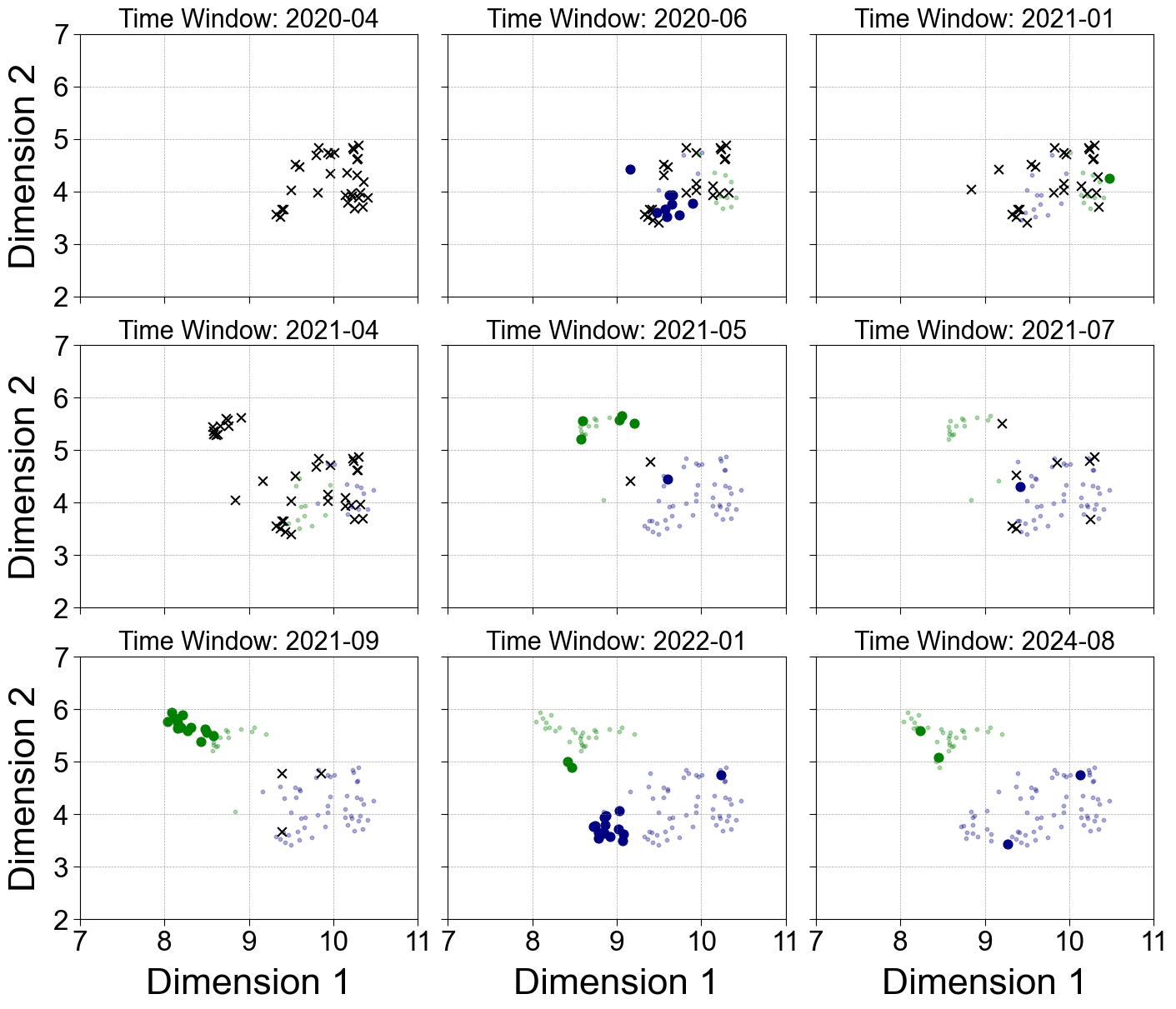}
    \caption{2D Scatter plot of the cumulative clustering obtained on \tp (after \texttt{UMAP} 10D reduction) over nine time windows, using ${\small\texttt{Solon-embeddings-large-0.1}}$. Outliers are indicated with black $\times$ and topics in blue and green.}
    \label{fig:solon-cinema-9tw}
\end{figure}

We applied topic-based clustering to the \tp dataset using the methodology described in subsection~\ref{ssec:methodology}. Figure~\ref{fig:solon-cinema-9tw} presents the cumulative clustering output generated by the ${\small\texttt{Solon-embeddings-large-0.1}}$ model. The figure shows a 2D representation derived from 10D \texttt{UMAP} projections of document embeddings across nine time windows, illustrating topic structure and outlier transitions over time.

Across all nine models and \texttt{UMAP} dimensions, clustering quality is consistent, with mean and median silhouette scores above 0.5 (range: -1 to 1). On average, body-text embeddings yield higher-quality clusters than headline or full-article representations. \texttt{UMAP} with 10 dimensions outperforms the 2D, 3D, and 5D settings. Among models, ${\small\texttt{Solon-embeddings-large-0.1}}$ achieves the highest scores, while ${\small\texttt{xlm-roberta-large}}$ performs the worst. Based on these findings, we evaluate Hypotheses~\h on \tp using \texttt{UMAP}-10D and body-text embeddings.

\begin{table}[h!]
  \caption{Mean silhouette scores per model for \texttt{UMAP} 10D using the body text of the \tp dataset. Bold values indicate the models achieving the best silhouette score for each document type. (See full results in \ref{sec:siloupilotappendix}.)}

  \label{tab:silhouette_scores_tp_10D}

  \centering%

  \resizebox{0.45\textwidth}{!}{%
    \begin{tabular}{lccc}
      \toprule
      \multirow{2}{*}{\textbf{Model}} & \multicolumn{3}{c}{\textbf{UMAP 10D}} \\
      \cmidrule(lr){2-4} 
                & \textbf{Headline} & \textbf{Body} & \textbf{Full Article} \\
      \midrule
      ${\texttt{multilingual-e5-large}}$          & \textbf{0.6065} & 0.5519 & 0.5689 \\
      ${\texttt{e5-base-v2}}$                     & 0.5592 & 0.5350 & 0.4846 \\
      ${\texttt{sentence-camembert-base}}$        & 0.5990 & 0.5850 & 0.6167 \\
      ${\texttt{all-MiniLM-L12-v2}}$              & 0.5654 & 0.5846 & 0.5349 \\
      ${\texttt{Solon-embeddings-large-0.1}}$     & 0.5772 & 0.6694 & 0.5553 \\
      ${\texttt{xlm-roberta-large}}$              & 0.4941 & 0.4802 & 0.4424 \\
      ${\texttt{all-roberta-large-v1}}$           & 0.5525 & 0.6258 & 0.5759 \\
      ${\texttt{multilingual-mpnet-base-v2}}$     & 0.5391 & 0.5923 & 0.6865 \\
      ${\texttt{distilbert-base-uncased}}$        & 0.3670 & \textbf{0.9373} & \textbf{0.8895} \\
      \midrule
      Mean                           & 0.5400 & \textbf{0.6180} & 0.5993 \\
      Median                         & 0.5417 & \textbf{0.6183} & 0.5756 \\
      \bottomrule
    \end{tabular}
  }
\end{table}

\subsection{Outlier Behavior}

\begin{figure}[h!]
    \centering
    \includegraphics[width=1\linewidth]{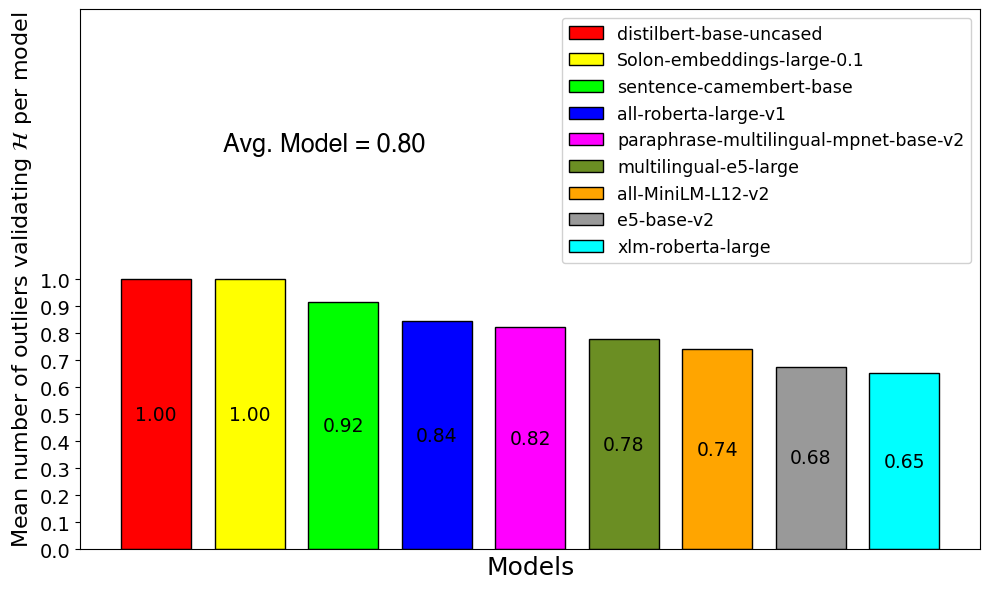}
    \caption{Mean number of outliers per model that validate prediction \h on \tp by converting into topic inlier at some time point (specific to each model). Each colored bar represents the mean of each model.}
    \label{fig:H_means_tp}
\end{figure}

To evaluate Hypothesis \h, we computed, for each model, the proportion of outliers that later became inliers during cumulative clustering. Figure~\ref{fig:H_means_tp} shows the mean validation score per model.

The average validation score of \h across models on \tp is high, with a mean of 0.80. As expected, models trained or fine-tuned on French perform strongly: ${\small\texttt{Solon-embeddings-large-0.1}}$ achieves perfect validation (1.0), and ${\small\texttt{sentence-camembert-base}}$ scores 0.92. Among English-language models, ${\small\texttt{e5-base-v2}}$ shows intermediate performance (0.68), while several others yield unexpectedly strong results: ${\small\texttt{all-MiniLM-L12-v2}}$ (0.74), ${\small\texttt{all-roberta-large-v1}}$ (0.84), and ${\small\texttt{distilbert-base-uncased}}$ (1.0). Multilingual models show mixed performance: ${\small\texttt{xlm-roberta-large}}$ scores moderately (0.65), whereas ${\small\texttt{paraphrase-multilingual-mpnet-base-v2}}$ (0.82) and ${\small\texttt{multilingual-e5-large}}$ (0.78) achieve high scores. Overall, model-level validation of \h ranges from moderate to perfect, with a relatively uniform distribution.

Across models, outlier-to-inlier conversion rates are highest in early clustering phases (64.58\%–100\% in late 2020), followed by a tapering trend with persistent outliers in later periods. As detailed in Appendix~\ref{sec:timepilotappendix}, some models exhibit stable integration while others decline over time. Despite these intra-model fluctuations, the general pattern of early integration supports \h across temporal windows.

To test whether this holds beyond model variation, we computed inter-agreement using the rescaling method mentioned in Section~\ref{ssec:outtopic}. The result \( a = 0.7002 \) shows strong agreement that \h is validated across all models based on converting the same outliers. This suggests that despite inconsistencies in how individual models integrate outliers over time, their validation of \h remains broadly aligned. Accordingly, the average model \(x=0.80\) is a good consensus model regarding the validation of \h. For this reason, we proceed using the average model representation for our interpretability analysis.

In the next section, we examine whether writing style, beyond semantic similarity, helps explain why some outliers are eventually integrated into clusters, while others remain isolated.

\subsection{Lexicon and Writing Style Analysis}

As an attempt to explain the conversion of outliers into topics, we controlled for topical alignment among outliers to assess their influence on topic formation (see Subsection~\ref{ssec:stylemethod}). For each word appearing in outlier documents, we computed the difference in average TF-IDF scores between those validating \h and those not validating \h. Specifically, we used the lexical salience metric \(\Delta \text{TFIDF}(w)\), as defined in~\eqref{eq:delta_tfidf}, and its inverse. Among the top 20 words in each class, the mean difference was $0.0088$ for outliers validating \h and $-0.0126$ for those not validating \h. Both differences were statistically significant at the $0.05$ level using the Kruskal–Wallis test.

A closer examination of the top 20 terms most prevalent among outliers validating \h reveals words associated with institutional support for the project (e.g., ``\textit{cabinet}'', ``\textit{total}'', ``\textit{lobbying}'', ``\textit{saclay}'') or individuals endorsing it (e.g., ``\textit{brunelle}'', ``\textit{nathalie}''). In contrast, the top 20 terms more prevalent among outliers \textit{not} validating \(\mathcal{H}\) include words reflecting opposition to the project (e.g., ``\textit{recours}'', ``\textit{victoire}''), as well as references to activist groups (e.g., ``\textit{greenpeace}'', ``\textit{militant}'') and opposing figures (e.g., ``\textit{julliard}'', ``\textit{jean}''). In both sets, the majority of these words were statistically distinctive.\footnote{Three words among outliers validating \h—``\textit{public}'', ``\textit{direction}'', and ``\textit{palaiseau}''—were not statistically significant, while only one word (``\textit{ecole}'') lacked significance among outliers not validating \h.} These results suggest that conversion of outliers into topics is partly influenced by their alignment with dominant themes in the \tp dataset, which is consistent with the role of semantic similarity in reinforcing or generating topical structure.

To evaluate our qualitative observation that the lexicon of outliers not validating \h tends to be more subjectively framed or opinion-laden, we carried out a quantitative analysis to test this hypothesis. Specifically, we assessed whether lexical salience defined in~\eqref{eq:delta_tfidf} correlated with the degree of subjectivity or neutrality in the documents where each word occurred. For each word \(w\), we computed the average subjectivity and neutrality scores across all documents \(D_w\) in which it appeared:

\begingroup
\small
\begin{align}
\text{Subjectivity}(w) &= \frac{1}{|D_w|} \sum_{d \in D_w} s(d) \\
\text{Neutrality}(w)   &= \frac{1}{|D_w|} \sum_{d \in D_w} n(d)
\end{align}
\endgroup

where \( s(d) \) and \( n(d) \) denote the subjectivity and neutrality scores of document \( d \), computed using \texttt{TextBlob}~\cite{loria2018textblob} and \texttt{VADER}~\cite{hutto2014vader}, respectively.

We then computed Spearman’s correlation between \(\Delta \text{TFIDF}(w)\) and the subjectivity and neutrality scores of the corresponding documents. The analysis revealed a \textit{moderate negative correlation} with subjectivity (\( r = -0.223 \), \( p < 0.01 \)) and a \textit{weak positive correlation} with neutrality (\( r = 0.105 \), \( p < 0.01 \)). These patterns indicate that words more prominent among converted outliers tend to appear in more neutral, less subjective contexts and thus that outliers more likely to become topics are characterized by a lexicon that is more factual in nature.

\begin{figure}[h!]
    \centering
    \begin{subfigure}[b]{\linewidth}
        \centering
        \includegraphics[width=\linewidth]{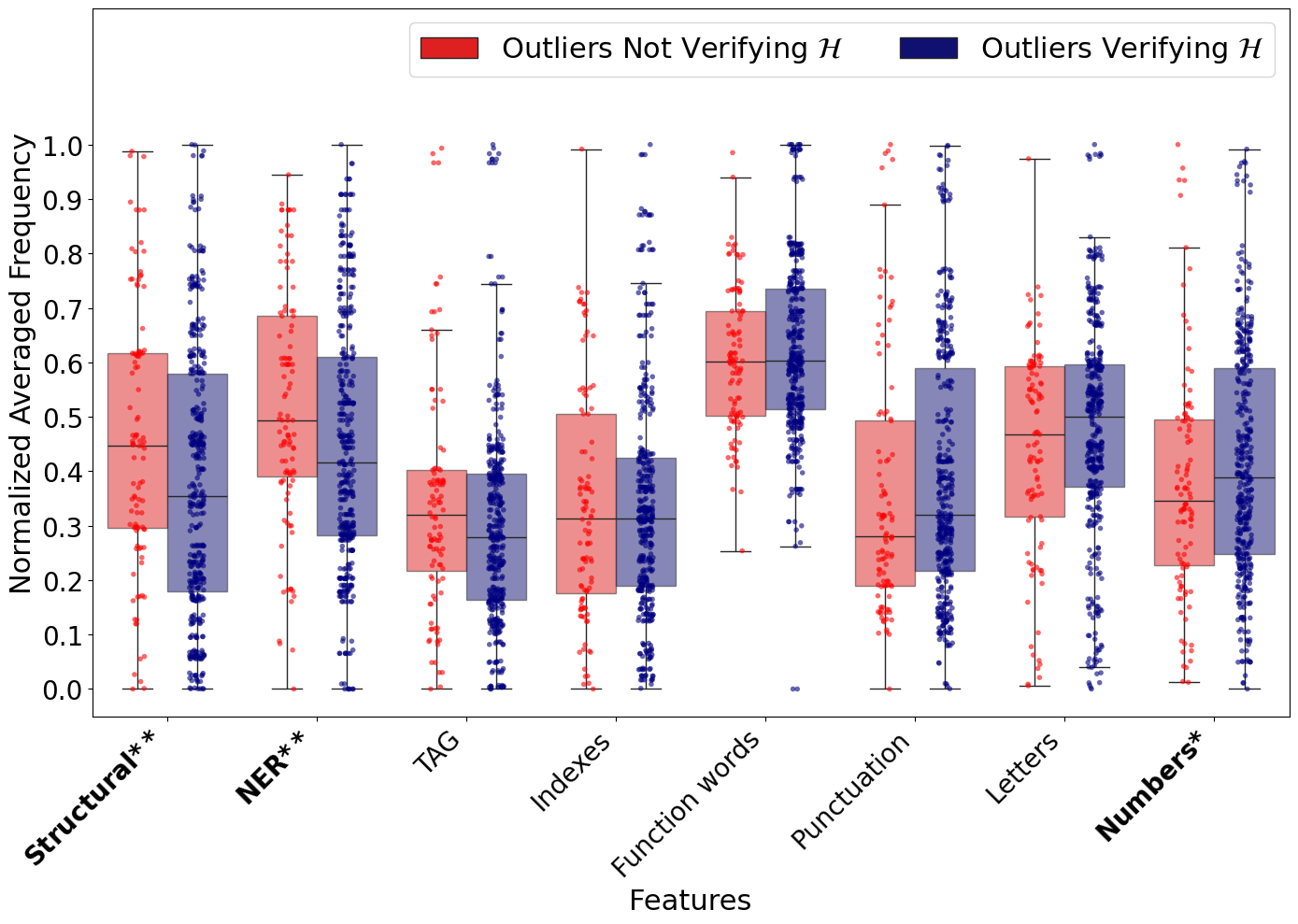}
        \caption{Differences in mean frequencies for the eight main features.}
        \label{fig:tp-models-boxplot-overall}
    \end{subfigure}
    
    \vspace{0.3cm}%
    
    \begin{subfigure}[b]{\linewidth}
        \centering
        \includegraphics[width=\linewidth]{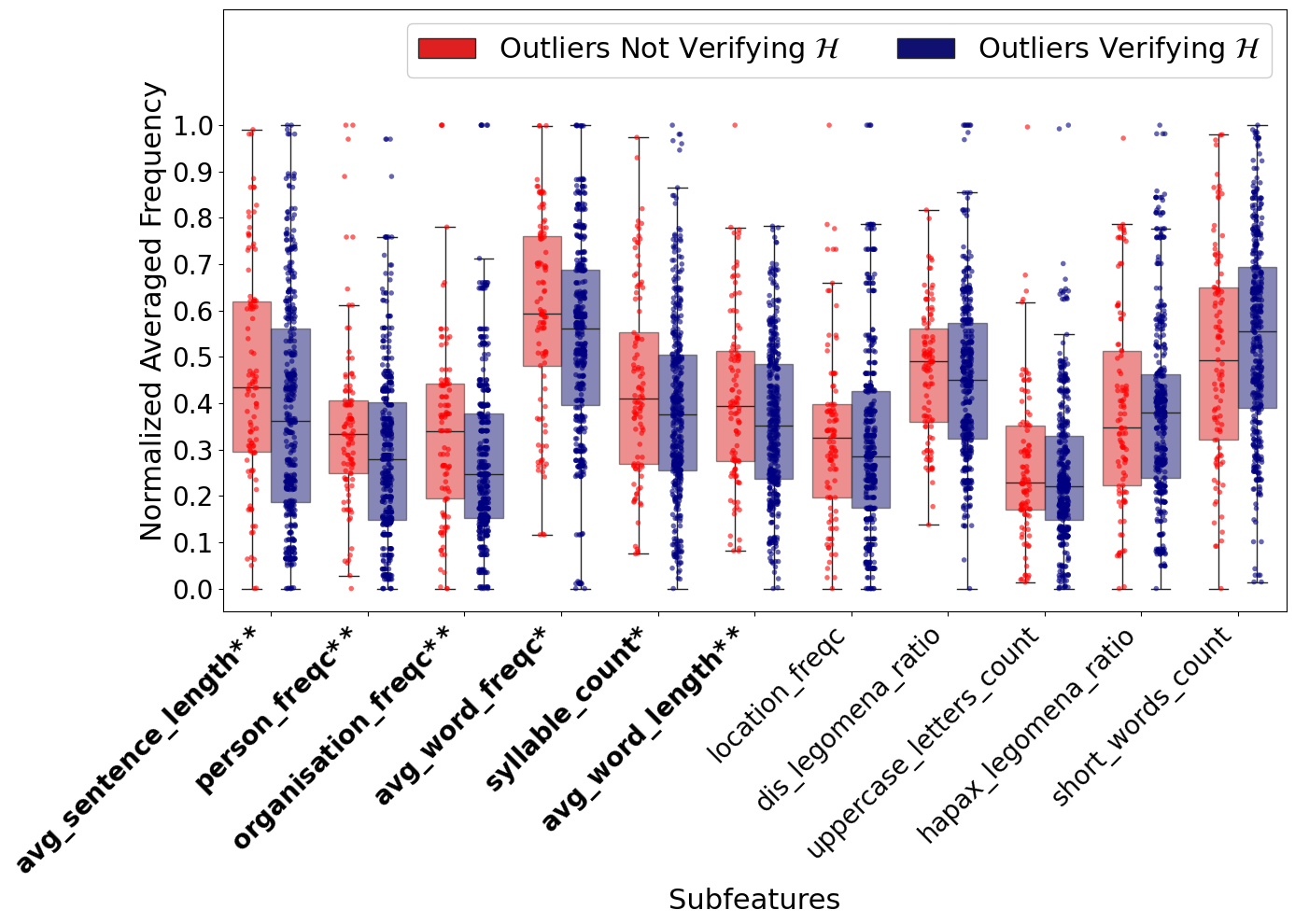}
        \caption{Differences in mean frequencies for subfeatures, based on observed significance in Figure \ref{fig:tp-models-boxplot-overall}.}
        \label{fig:tp-models-boxplot-detailed}
    \end{subfigure}
    
    \caption{Differences for \tp in the eight stylistic features and subfeatures from \citet{terreau2021writing}, between outliers validating \h and outliers not validating \h.  Statistical significance is measured using the Kruskal–Wallis test, with $^{*}$ and $^{**}$ indicating $p$ values $< 0.05$ and $< 0.01$, respectively.}
    \label{fig:tp-models-boxplot}
\end{figure}

To evaluate broader stylistic effects, we applied the stylometric framework of \citet{terreau2021writing} to measure differences across eight core stylistic features between converted and non-converted outliers: \textit{function words}, \textit{punctuation marks}, \textit{numbers}, \textit{named entities}, \textit{part-of-speech tags}, \textit{structural features}, \textit{lexical complexity indexes}, and \textit{readability metrics}. Figure~\ref{fig:tp-models-boxplot} summarizes the results for both main categories (Fig.~\ref{fig:tp-models-boxplot-overall}) and significant subfeatures (Fig.~\ref{fig:tp-models-boxplot-detailed}). We omit a detailed analysis for significant differences in Numbers, as this feature consists solely of single-digit values (ranging from 0 to 9), making any further breakdown not directly interpretable. 

At the level of the eight main features (see Figure~\ref{fig:tp-models-boxplot-overall}), statistically significant differences were observed only for Named Entities (NER), Structural features, and Numbers. Structural features and NER were less frequent in outliers validating \h than in those not validating \h, whereas Numbers were more frequent in outliers validating \h. No significant differences were found for TAG, Punctuation, Letters, Indexes, or Function Words.

A closer examination of the significant subfeatures (Fig.~\ref{fig:tp-models-boxplot-detailed}) shows that, for NER, names of persons and organizations appear significantly less often in outliers validating \h. No difference was found for location markers. For Structural subfeatures, outliers validating \h exhibit shorter sentences and words, fewer syllables per word, and higher average word frequency. No other structural subfeatures showed significant variation.

These stylistic differences observed for the average model may be explained by the fact that more structural features introduce complexity, and thus stylistic simplification may support the integration of outliers into topic clusters. Specifically, shorter and simpler text, with fewer named entities, may make it easier for the average model to associate such outliers with broader topic structures, thus facilitating the validation of \h. Conversely, a higher frequency of Numbers, particularly single-digit ones, may reflect more patterned or categorical language that also facilitates topic clustering. No clear effects were found for TAG, Punctuation, Letters, Indexes, or the remaining structural subfeatures.

\section{Replication Study}
\label{sec:replication}

\subsection{English Dataset}

To validate and generalize our findings, we used an existing larger English dataset of climate change news articles, \textit{climate-news-db}.\footnote{\url{https://www.climate-news-db.com}} This dataset originally comprised 27,877 news articles from global media outlets, spanning January 2015 to November 2024. To ensure topical consistency, we curated a focused subset of 312 articles, referred to as \ghg, by filtering for content explicitly addressing Greenhouse Gas Emissions (GHG). Articles were selected based on the presence of the terms ``\textit{Greenhouse Gas}'' or ``\textit{Greenhouse Emissions}'', and sampled across 20 monthly time windows between January 2022 and August 2023. For consistency, we retained only articles from major U.S.-based outlets (e.g., \textit{The Washington Post}, \textit{The New York Times}, \textit{Fox News}, and \textit{CNN}).

\subsection{Topic-Based Clustering}

We applied topic-based clustering to the body text of the \ghg articles using 10D \texttt{UMAP} projections. With the exception of ${\small\texttt{e5-base-v2}}$, Table~\ref{tab:umap_10d_body_ghg} shows that all nine models achieved strong silhouette scores, with both mean and median values at or above 0.5 (on a scale from –1 to 1). These results are slightly lower than, but broadly consistent with, those obtained for the \tp dataset under the same configuration.

\begin{figure}[h!]
    \centering
    \includegraphics[width=1\linewidth]{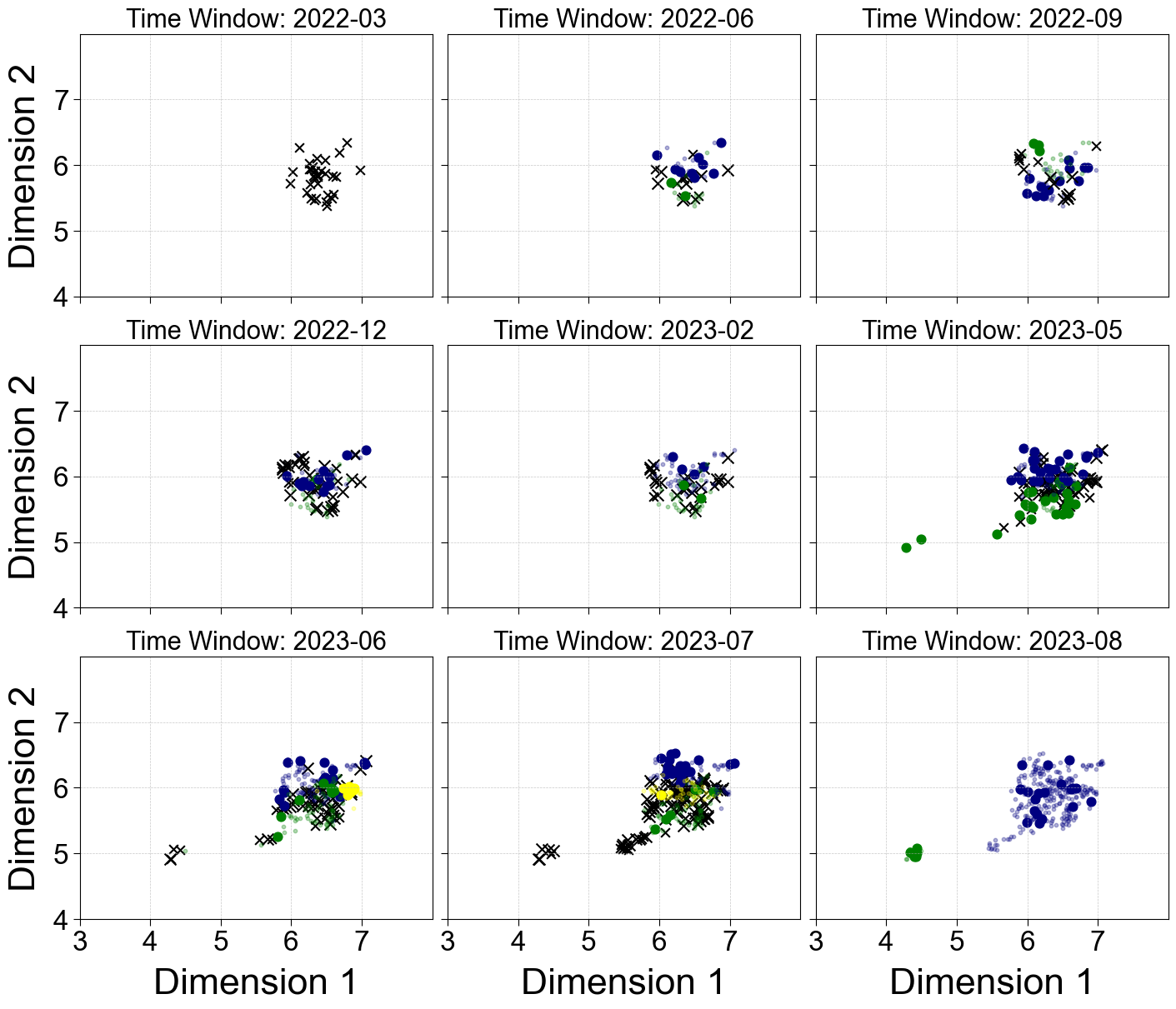}
    \caption{2D Scatter plot of the \texttt{UMAP} 10D cumulative clustering obtained on \ghg over nine time windows, using ${\small\texttt{e5-base-v2}}$. Outliers are indicated with black $\times$, topics in blue, green and yellow.}
    \label{fig:e5-cinema-9tw}
\end{figure}

\begin{table}[h!]
  \caption{\texttt{UMAP} 10D silhouette scores obtained on the \ghg dataset for the body text of articles, sorted from best to worst.\label{tab:umap_10d_body_ghg}}
  \scriptsize%
  \centering%
  \resizebox{0.45\textwidth}{!}{%
    \begin{tabular}{lc}
      \toprule
      \textbf{Model} & \textbf{Mean Silhouette Score} \\
      \midrule
      ${\texttt{e5-base-v2}}$ & \textbf{0.5661}\\
      ${\texttt{multilingual-e5-large}}$ & 0.5490\\
      ${\texttt{all-MiniLM-L12-v2}}$ & 0.5416\\
      ${\texttt{...-multi..-mpnet-base-v2}}$ & 0.5387\\
      ${\texttt{xlm-roberta-large}}$ & 0.5376\\
      ${\texttt{Solon-embeddings-large-0.1}}$ & 0.5159\\
      ${\texttt{sentence-camembert-base}}$ & 0.5092\\
      ${\texttt{all-roberta-large-v1}}$ & 0.5044\\
      ${\texttt{distilbert-base-uncased}}$ & 0.4998\\
      \midrule
      Mean & 0.5291\\
      Median & 0.5376\\
      \bottomrule
    \end{tabular}}

\end{table}

\subsection{Outlier Behavior}

\begin{figure}[htpb]
    \centering
    \includegraphics[width=1\linewidth]{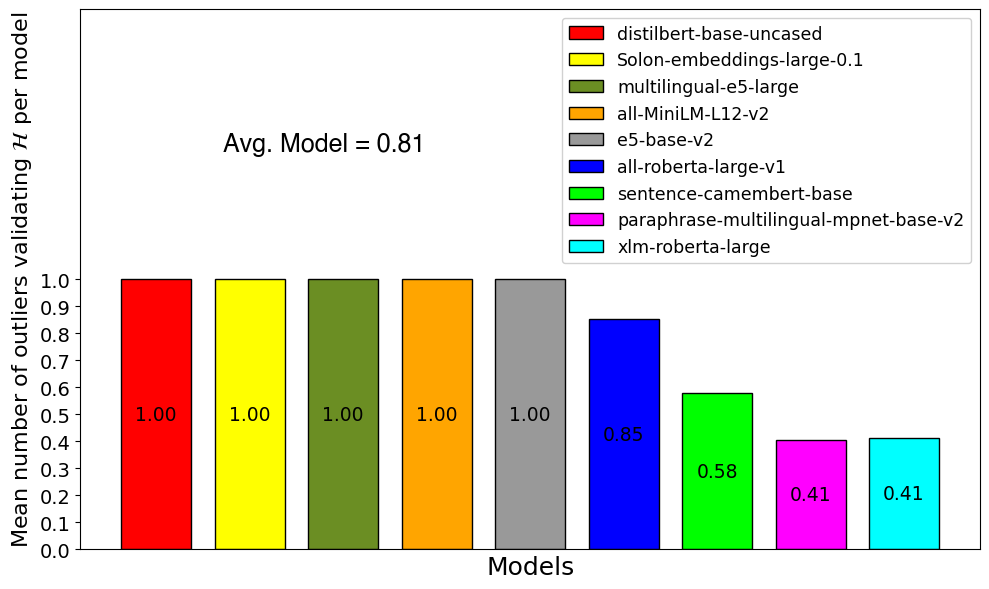}
    \caption{Mean number of outliers per model that validate prediction \h on \ghg by converting into topic inlier at some time point (specific to each model). Each colored bar represents the mean for each model.}
    \label{fig:H_means_ghg}
\end{figure}

Figure~\ref{fig:H_means_ghg} shows the mean validation score per model for \h on \ghg. The results indicate a high average validation across models, with a mean score of 0.81. As expected, English-specialized models:   ${\small\texttt{distilbert-base-uncased}}$, ${\small\texttt{e5-base-v2}}$, and ${\small\texttt{all-MiniLM-L12-v2}}$, achieve perfect validation (1.0), followed by ${\small\texttt{all-roberta-large-v1}}$ (0.85). Among French-specialized models, ${\small\texttt{sentence-camembert-base}}$ performs more weakly (0.58), as anticipated, while the perfect score of ${\small\texttt{Solon-embeddings-large-0.1}}$ (1.0) is less expected. Multilingual models show mixed re\-sults: ${\small\texttt{paraphrase-multilingual-mpnet-base-v2}}$ and ${\small\texttt{xlm-roberta-large}}$ perform poorly (both 0.41), while ${\small\texttt{multilingual-e5-large}}$ again achieves perfect validation. The distribution of scores appears bimodal: five models achieve perfect validation, while the remaining four show moderate to low scores. This sharp divide may reflect potential overfitting among English-specialized models that integrate all outliers into topics.

This consistency in temporal dynamics (see Appendix~\ref{sec:timereplicappendix} for a detailed time-window analysis) aligns with the high average validation score of 0.81 (Figure~\ref{fig:H_means_ghg}). Most models follow a similar pattern: strong early outlier-to-topic conversion, reduced integration in mid-phases, and stabilization with persistent outliers. While some models, particularly multilingual ones and \texttt{sentence-camembert-base}, show greater fluctuation, the overall trend supports \h. As in the Pilot experiment, we computed inter-agreement across models with respect to \h, using the rescaling method of \citet{icard2024multi}. Again, the result \(a=0.6783\) strongly supports that models validate \(\mathcal{H}\) based on converting the exact same outliers. The average model \(x=0.81\) is then a good consensus model regarding the validation of \h.

\begin{figure}[htpb]
    \centering
    \includegraphics[width=1\linewidth]{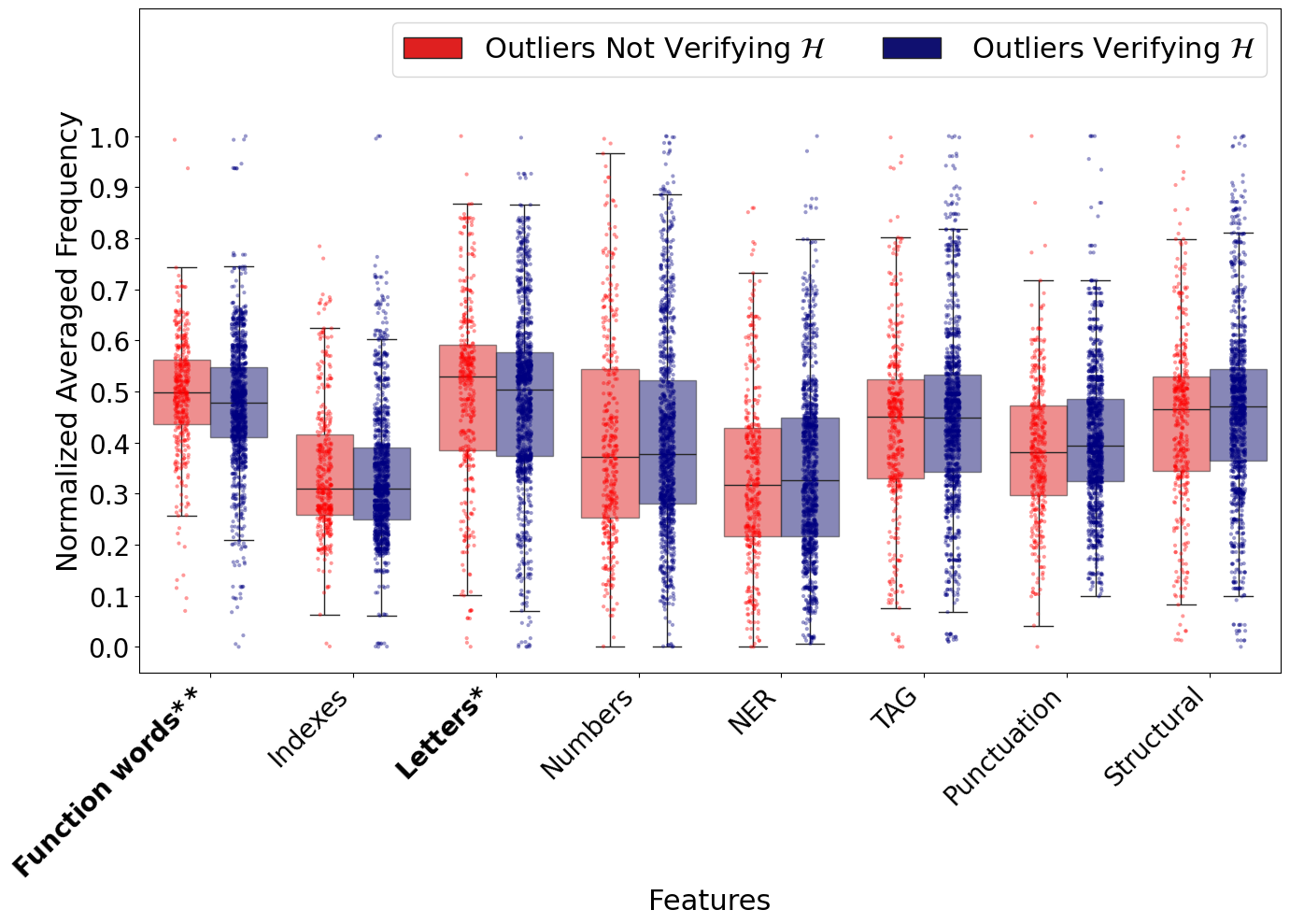}
   \caption{Differences for \ghg in the eight stylistic features from \citet{terreau2021writing}, between outliers validating \h and outliers not validating \h.  Statistical significance is measured using the Kruskal–Wallis test, with $^{*}$ and $^{**}$ indicating $p$ values $< 0.05$ and $< 0.01$, respectively.}
    \label{fig:ghg-models-boxplot}
\end{figure}

\subsection{Lexicon and Writing Style Analysis}

As part of our interpretability analysis, we sought to understand why some outliers aligned with topics while others did not. We first examined the top 20 words with the highest \(\Delta \text{TFIDF}\) scores in outliers validating \(\mathcal{H}\) compared to those not validating it, and vice versa. As defined in~\eqref{eq:delta_tfidf}, \(\Delta \text{TFIDF}(w)\) captures the difference in average TFIDF scores for word \(w\) between the two outlier classes. The mean difference was 0.0031 for~\eqref{eq:delta_tfidf}, and 0.0023 for the reverse. Neither difference was statistically significant (Kruskal--Wallis test, \(p > 0.05\)), suggesting that thematic lexical content does not meaningfully distinguish the two outlier classes in \ghg{}. However, this finding does not rule out the possibility that stylistic or other non-topical lexical and linguistic features influence outlier conversion.

To address this gap, we analyzed the differences in stylistic characteristics between the two outlier classes, using the framework proposed by \citet{terreau2021writing}. The results for \ghg are given in Figure \ref{fig:ghg-models-boxplot} for the eight main features. We do not provide a detailed analysis of Function words and Letters, as Letters consist solely of single-character values (ranging from A to Z), and Function Words gain significance from their overall distribution rather than their individual occurrences, making a further breakdown not directly interpretable.

Among the eight features, significant frequency differences were found only for Function words and Letters, which were notably less frequent in outliers verifying \h compared to outliers not verifying \h. This may be explained by the fact that function words (e.g., prepositions, conjunctions) and letters (e.g., A, B, C) lack semantic content, so their reduction helps the average model recognize topics in outliers and validate \h. In contrast, Indexes, Numbers, NER, Punctuation, TAG, and Structural features do not appear to have a particular effect on this recognition.

\section{Discussion}
\label{sec:discussion}

We observed consistent outlier-to-topic conversion across two linguistically distinct datasets, confirming that the phenomenon generalizes. Validation of \h is robust across topic domains (social responsibility and climate change), languages (French and English), and dataset sizes (102 and 312 articles), with a stable mean score around 0.80. Inter-model agreement remains high (with \(a=0.7002 \) for French, \(a=0.6783\) for English), suggesting that topic-based clustering reliably integrates outliers under varied conditions.

In lexical analysis, TF-IDF differences between converted and non-converted outliers were significant in \tp but not in \ghg. In \tp, converted outliers were more strongly associated with lower subjectivity and higher lexical neutrality. This reflects a structural difference: \tp focuses on a defined controversy with a polarized lexicon, while \ghg likely follows a more neutral, report-oriented style, as it was not curated under controversy criteria.

The stylistic features analysis revealed that writing style has a significant impact on the conversion of outliers into topics, though the relevant features differ by language. In \tp, conversion is influenced by structural features, named entities, and numbers; in \ghg, by function words and letter distributions. This suggests that embedding models rely on language-specific stylistic cues when integrating outliers.

These differences align with model training: French-trained models perform better on \tp, English-trained ones on \ghg, while multilingual models show mixed results, reflecting their training data (see Table \ref{tab:detailmodels} in Appendix \ref{sec:timereplicappendix} for details).

\section{Conclusion}
\label{sec:conclusion}

Our findings demonstrate that outlier-to-inlier conversion is a consistent mechanism in topic emergence within cumulative, density-based clustering frameworks. The effect is robust across nine language models, two typologically distinct languages, and datasets with varying topical scope. In the French dataset (\tp), focused on a well-defined controversy, average model validation reached 0.80; in the English dataset (\ghg), covering broader climate discourse, the score was similarly high at 0.81. Inter-model agreement exceeded 0.65 in both cases, indicating stable clustering dynamics across architectures and domains.

Future work will distinguish between outliers that act as precursors to new topics and those that reinforce existing structures. We aim to quantify their predictive value and examine their temporal behavior across phases of topic development.

We also plan to scale our analysis to larger and more heterogeneous corpora, particularly in domains where informational risks, such as discursive conflict and disinformation, are likely to emerge or escalate.
In parallel, we will evaluate alternative clustering algorithms with integrated outlier detection (e.g., \texttt{OPTICS}) and broaden our assessment across additional model architectures.
These extensions aim to test the generality and deepen the explanatory power of our findings.

\section*{Limitations} 

This study was designed as a controlled pilot to explore the predictive role of outliers in topic emergence under well-defined experimental conditions. Although the number of raw articles was relatively limited (102 in French and 312 in English), each document was processed with nine distinct language models, resulting in 918 French and 2,808 English data points. This mitigated the limitations typically associated with small corpus sizes. 

High inter-model agreement (\(a=0.7002 \) for French and \(a=0.6783\) for English) and consistent clustering quality (silhouette scores of 0.61 and 0.52, respectively) further support that the results are robust within the bounds of this setup. 

The decision to prioritize depth over breadth at the expense of dataset size was deliberate: it enabled the construction of a high-quality, manually curated corpus with full-text availability, temporal continuity (i.e., no temporal gaps), and source diversity. This design helped control for confounding factors such as incomplete timelines and uneven topic coverage, which often affect large-scale datasets whose compilation processes are not fully transparent.

While these constraints were necessary to ensure experimental clarity and interpretability, they naturally limit the generalizability of the findings. Future work will scale the analysis to larger corpus of news articles to test its applicability in more complex and dynamic information environments.

\section*{Ethics Statement}

Our research adheres to the ethical principles of open science, transparency, and sustainability. We ensure reproducibility by making our code accessible in a dedicated \href{https://github.com/evangeliazve/outliers-to-topics-icnlsp}{GitHub repository}, with data and results available upon request. We comply with intellectual property and data protection regulations by sharing only vector embeddings generated by language models. This approach aligns with the principles of `transformative fair use''. We promote AI transparency by contributing to the interpretability of language models, supporting the responsible and explainable use of these models. To support efficiency and sustainability, we prioritize the use of small, open-source language models.

\section*{Declaration of contribution}
EZ, BI, and JGG conceptualized the research problem and designed the experiments. EZ managed the data collection process. AB and LS managed data cleaning and annotation. EZ was responsible for the technical aspects: coding, model selection, and building the experimental framework. EZ, BI, GB, and JGG analyzed and discussed the results. EZ and BI wrote the paper, which all authors read and revised together. EZ and BI share first authorship.
Correspondence: evangelia.zve@lip6.fr, benjamin.icard@lip6.fr, jean-gabriel.ganascia@lip6.fr.

\bibliographystyle{acl_natbib}
\bibliography{sample-base}

\clearpage%
\appendix
\section{Appendix}
\label{sec:appendix}

\setlength\dashlinedash{0.6pt}
\setlength\dashlinegap{2pt}
\begin{table*}[t]
\centering
\normalsize
\label{tab:modelplot}
\hspace*{0.05\textwidth}%
\resizebox{0.7\textwidth}{!}{%
\begin{tabular}{@{}lcccc@{}}
\toprule
\textbf{Model} & \textbf{Architecture} & \textbf{Dimensions} & \textbf{Language} & \textbf{Parameters} \\
\midrule
\normalsize{\texttt{Solon-embeddings-large-0.1}} & RoBERTa & 1024 & \multirow{2}{*}{French} & 560M \\
\normalsize{\texttt{sentence-camembert-base}} & CamemBERT & 768 &  & 111M \\
\cdashline{1-5}
\normalsize{\texttt{all-roberta-large-v1}} & RoBERTa & 1024 & \multirow{4}{*}{English} & 355M \\
\normalsize{\texttt{e5-base-v2}} & E5 & 768 &  & 109M \\
\normalsize{\texttt{distilbert-base-uncased}} & DistilBERT & 768 &  & \ \ 67M \\
\normalsize{\texttt{all-MiniLM-L12-v2}} & MiniLM & 384 &  & 33.4M \\
\cdashline{1-5}
\normalsize{\texttt{xlm-roberta-large}} & XLM-RoBERTa & 1024 & \multirow{3}{*}{Multilingual} & 561M \\
\normalsize{\texttt{multilingual-e5-large}} & E5 & 1024 &  & 560M \\
\normalsize{\texttt{paraphrase-multilingual-mpnet-base-v2}} & MPNet & 768 &  & 278M \\
\bottomrule
\end{tabular}}
\hspace*{0.05\textwidth}%
\caption{Description of the nine sentence embedding models used to conduct the topic-based clustering experiments.}
\label{tab:detailmodels}
\end{table*}

\subsection{Supplementary Materials}
The code and visualizations supporting this paper are available at: \url{https://github.com/evangeliazve/outliers-to-topics-icnlsp}. The datasets and experimental results can be provided upon request. The repository includes Python scripts for reproducing our experiments, as well as statistical analyses and visualizations corresponding to key figures and tables in the paper.
The \texttt{BERTopic} framework is documented at: \url{https://maartengr.github.io/BERTopic/}.
Further details on \texttt{HDBSCAN} can be found in its official documentation: \url{https://hdbscan.readthedocs.io/en/latest/}, and information on UMAP dimensionality reduction is available at: \url{https://umap-learn.readthedocs.io/en/latest/basic_usage.html}.
For TF-IDF, we used the \texttt{TfidfVectorizer} from scikit-learn: \url{https://scikit-learn.org/stable/modules/generated/sklearn.feature_extraction.text.TfidfVectorizer.html}.
The sentiment analysis tools employed in this study are \texttt{TextBlob} (\url{https://textblob.readthedocs.io/en/dev/index.html}) and \texttt{VADER} (\url{https://github.com/cjhutto/vaderSentiment}).

\subsection{Models}
\label{sec:modelappendix}

Table~\ref{tab:detailmodels} presents the nine sentence embedding models used in our experiments for topic-based clustering, detailing their underlying architectures, embedding dimensionality, language coverage, and model sizes.

\subsection{Pilot Study Appendix}
\label{sec:pilotappendix}

\subsubsection{Detailed Silhouette Scores}
\label{sec:siloupilotappendix}

In this appendix, we provide detailed results from the pilot study evaluating the effectiveness of different sentence embedding models for topic-based clustering on the \tp dataset. Specifically, Table~\ref{tab:silhouette_scores_tp} reports the mean silhouette scores obtained for each model under varying dimensionality reductions (2D, 3D, 5D, and 10D using \texttt{UMAP}) and different text sample types (headline, body, and combined text). These results offer insights into how model selection, dimensionality, and text granularity impact clustering quality.

\begin{table*}[t]
  \tiny
  \resizebox{\textwidth}{!}{%
    \begin{tabular}{lcccccccccccc}
      \toprule
      \multirow{2}{*}{\textbf{Model}} & \multicolumn{3}{c}{\textbf{UMAP 2D}} & \multicolumn{3}{c}{\textbf{UMAP 3D}} & \multicolumn{3}{c}{\textbf{UMAP 5D}} & \multicolumn{3}{c}{\textbf{UMAP 10D}} \\
      \cmidrule(lr){2-4} \cmidrule(lr){5-7} \cmidrule(lr){8-10} \cmidrule(lr){11-13}
                & \textbf{Headline} & \textbf{Body} & \textbf{All} & \textbf{Headline} & \textbf{Body} & \textbf{All} & \textbf{Headline} & \textbf{Body} & \textbf{All} & \textbf{Headline} & \textbf{Body} & \textbf{All} \\
      \midrule
      ${\texttt{multilingual-e5-large}}$          & 0.6235 & 0.6002 & 0.5914 & 0.6121 & 0.5480 & 0.5713 & 0.6020 & 0.5481 & 0.5692 & 0.6065 & 0.5519 & 0.5689 \\
      ${\texttt{e5-base-v2}}$                     & 0.6133 & 0.5556 & 0.4668 & 0.5718 & 0.5627 & 0.4671 & 0.5580 & 0.5479 & 0.5030 & 0.5592 & 0.5350 & 0.4846 \\
      ${\texttt{sentence-camembert-base}}$        & 0.6120 & 0.5616 & 0.5994 & 0.6083 & 0.5791 & 0.6302 & 0.5934 & 0.5877 & 0.6354 & 0.5990 & 0.5850 & 0.6167 \\
      ${\texttt{all-MiniLM-L12-v2}}$              & 0.6039 & 0.5858 & 0.5465 & 0.5570 & 0.6197 & 0.5243 & 0.5702 & 0.4962 & 0.5608 & 0.5654 & 0.5846 & 0.5349 \\
      ${\texttt{Solon-embeddings-large-0.1}}$     & 0.5573 & 0.6340 & 0.6497 & 0.6056 & 0.6351 & 0.6031 & 0.5660 & 0.6153 & 0.5778 & 0.5772 & 0.6694 & 0.5553 \\
      ${\texttt{xlm-roberta-large}}$              & 0.5416 & 0.3729 & 0.3294 & 0.4996 & 0.3812 & 0.3226 & 0.5348 & 0.3694 & 0.3848 & 0.4941 & 0.4802 & 0.4424 \\
      ${\texttt{all-roberta-large-v1}}$           & 0.5294 & 0.6701 & 0.5862 & 0.5427 & 0.6255 & 0.6121 & 0.5536 & 0.6361 & 0.6040 & 0.5525 & 0.6258 & 0.5759 \\
      ${\texttt{..-multilingual-mpnet-base-v2}}$ & 0.5259 & 0.6062 & 0.7324 & 0.5221 & 0.5918 & 0.6429 & 0.5517 & 0.5977 & 0.6754 & 0.5391 & 0.5923 & 0.6865 \\
      ${\texttt{distilbert-base-uncased}}$        & 0.4872 & 0.7907 & 0.8535 & 0.5232 & 0.9233 & 0.8509 & 0.4413 & 0.9575 & 0.8670 & 0.3670 & 0.9373 & 0.8895 \\
      \midrule
      Mean                           & 0.5660 & 0.5975 & 0.5945 & 0.5603 & 0.6074 & 0.5816 & 0.5523 & 0.5951 & 0.6008 & 0.5400 & 0.6180 & 0.5993 \\
      Median                         & 0.5588 & 0.6056 & 0.5929 & 0.5692 & 0.5984 & 0.5718 & 0.5544 & 0.6029 & 0.6040 & 0.5417 & 0.6183 & 0.5756 \\
      \bottomrule
    \end{tabular}
  }
    \caption{Mean silhouette scores per model, dimensionality and text samples types obtained on dataset \tp.}
    \label{tab:silhouette_scores_tp}
\end{table*}

\subsubsection{Validation or invalidation of \h per model over different time windows for TP}
\label{sec:timepilotappendix}

For a more detailed examination of model variations, both across and within models, Table \ref{tab:timeconversiontp} presents the validation or invalidation of \h per model over different periods of cumulative clustering. Among French models, ${\small\texttt{Solon-embeddings-large-0.1}}$ is fully consistent, achieving complete integration early, while ${\small\texttt{sentence-camembert-base}}$ shows non-monotony, with conversion dropping from 95.83\% to 50\% and some persistent outliers. English models exhibit pronounced inconsistency: ${\small\texttt{e5-base-v2}}$ weakens over time (68.75\% to 36.67\%), ${\small\texttt{all-MiniLM-L12-v2}}$ and ${\small\texttt{all-roberta-large-v1}}$ show fluctuating progress despite strong early conversion (66.67\% and 85.42\%), and ${\small\texttt{distilbert-base-uncased}}$ remains fully stable with no outliers through the whole. Multilingual models vary widely, with ${\small\texttt{paraphrase-multilingual-mpnet-base-v2}}$ and ${\small\texttt{xlm-roberta-large}}$ starting strong (85.42\%, 64.58\%) but leaving substantial outliers later (12.74\%, 42.15\%), while ${\small\texttt{multilingual-e5-large}}$ follows an unstable trajectory, declining from 83.33\% to 48.15\%.

\begin{table*}[t]
\centering
\resizebox{\textwidth}{!}{
\begin{tabular}{llcccc}
\toprule
\multirow{2}{*}{\textbf{Model}} & \multirow{2}{*}{\textbf{Measures}} & \multicolumn{4}{c}{\textbf{Time}} \\
\cmidrule{3-6}
& & \textbf{2020-11 (50\%)}& \textbf{2021-07 (70\%)}& \textbf{2023-09 (90\%)}& \textbf{Remaining (100\%)} \\
\midrule
\multirow{3}{*}{${\texttt{Solon-..-large-0.1}}$} 
& Nb Outliers / All Articles at $t$ & 48/48& 8/69& 0/100& 0/102\\
& \% Becoming Inliers at $(t+n)$ & 100\% & 100\% & 0.00 & Converted on 2022-01 \\
\midrule
\multirow{3}{*}{${\texttt{..-multi..-mpnet-..}}$} 
& Nb Outliers / All Articles at $t$ & 48/48& 69/69& 15/100& 13/102\\
& \% Becoming Inliers at $(t+n)$ & 85.42\% & 84.06\% & 46.47\% & - \\
\midrule
\multirow{3}{*}{${\texttt{sentence-camembert-..}}$} 
& Nb Outliers / All Articles at $t$ & 48/48& 33/69& 8/100& 4/102\\
& \% Becoming Inliers at $(t+n)$ & 95.83\% & 90.91\% & 50.00\% & - \\
\midrule
\multirow{3}{*}{${\texttt{multi..-e5-large}}$} 
& Nb Outliers / All Articles at $t$ & 48/48& 25/69& 27/100& 25/102\\
& \% Becoming Inliers at $(t+n)$ & 83.33\% & 64.00\% & 48.15\% & - \\
\midrule
\multirow{3}{*}{${\texttt{xlm-roberta-large}}$} 
& Nb Outliers / All Articles at $t$ & 48/48& 39/69& 30/100& 43/102\\
& \% Becoming Inliers at $(t+n)$ & 64.58\% & 64.10\% & 26.67\% & - \\
\midrule
\multirow{3}{*}{${\texttt{all-MiniLM-L12-v2}}$} 
& Nb Outliers / All Articles at $t$ & 48/48& 69/69& 16/100& 26/102\\
& \% Becoming Inliers at $(t+n)$ & 66.67\% & 73.91\% & 12.50\% & - \\
\midrule
\multirow{3}{*}{${\texttt{all-roberta-large-v1}}$} 
& Nb Outliers / All Articles at $t$ & 48/48& 21/69& 10/100& 12/102\\
& \% Becoming Inliers at $(t+n)$ & 85.42\% & 71.43\% & 30.00\% & - \\
\midrule
\multirow{3}{*}{${\texttt{distil..-base-uncased}}$} 
& Nb Outliers / All Articles at $t$ & 0/48& 0/69& 0/100& 0/102\\
& \% Becoming Inliers at $(t+n)$ & 0.00\% & 0.00\% & 0.00\% & Converted on 2020-06 \\
\midrule
\multirow{3}{*}{${\texttt{e5-base-v2}}$} 
& Nb Outliers / All Articles at $t$ & 48/48& 21/69& 30/100& 41/102\\
& \% Becoming Inliers at $(t+n)$ & 68.75\% & 42.86\% & 36.67\% & - \\
\bottomrule
\end{tabular}
}
\caption{Proportion of outliers converting to clusters in \tp, for each model and along four time windows.}
\label{tab:timeconversiontp}
\end{table*}

Across models, a general pattern emerges: strong early conversion of outliers into topic inliers, slowing integration in the mid-phase, and eventual stabilization with persistent outliers in 2023. Early clustering is largely consistent, with conversion rates ranging from 64.58\% (${\texttt{xlm-roberta-large}}$) to 100\% (${\texttt{Solon-embeddings-large}}{\texttt{-0.1}}$) in 2020-11. By the mid-phase (2021-07), some models, like ${\texttt{all-MiniLM-L12-v2}}$ (73.91\%) and ${\texttt{all-roberta-large-v1}}$ (71.43\%), sustain moderate integration, while others, like ${\texttt{e5-base-v2}}$ (42.86\%), decline. Late-stage variations are more pronounced, with ${\texttt{paraphrase-}}{\texttt{multilingual-mpnet-base-v2}}$ retaining 46.47\% of outliers as topic inliers, while ${\texttt{xlm-roberta-large}}$ and ${\texttt{e5-base-v2}}$ drop to 26.67\% and 36.67\%, respectively. ${\texttt{sentence-camembert-base}}$, despite an early peak (95.83\%), declines to 50.00\%.  

\subsubsection{Top 10 Distinguishing Terms Based on TF-IDF Differences Between Outliers Validating and Not Validating \h}

Table~\ref{tab:tfidf-tp} lists the top 10 terms whose TF-IDF scores most strongly differentiate outliers that validate hypothesis \h from those that do not, highlighting key lexical features associated with each group.

\renewcommand{\arraystretch}{1.15}
\begin{table*}[t]
\centering
\resizebox{0.7\textwidth}{!}{%
\begin{tabular}{|l|r|r||l|r|r|}
\hline
\textbf{Word} &  $\boldsymbol{\Delta \mathrm{TFIDF}(w)}$ & $\boldsymbol{\Delta \mathrm{Occ}(w)}$ &
\textbf{Word} &  $\boldsymbol{\Delta \mathrm{TFIDF}(w)}$ & $\boldsymbol{\Delta \mathrm{Occ}(w)}$ \\
\hline
cabinet     &  $0.0122^{*}$   &   93  & totalenergies & $-0.0328^{**}$ & -28  \\
\hline
total       &  $0.0119^{*}$   & 2613  & recours       & $-0.0185^{**}$ & -14  \\
\hline
brunelle    &  $0.0106^{*}$   &  136  & greenpeace    & $-0.0173^{**}$ & -265 \\
\hline
nathalie    &  $0.0104^{*}$   &  139  & victoire      & $-0.0155^{**}$ & -6   \\
\hline
lobbying    &  $0.0103^{**}$  &  122  & ecole         &       $-0.0143$      & -162 \\
\hline
public      &  $0.0098$       &  428  & julliard      & $-0.0129^{**}$ & -14  \\
\hline
direction   &  $0.0097$       &  563  & jean          & $-0.0126^{**}$ & -20  \\
\hline
palaiseau   &  $0.0095$       &   60  & décision      & $-0.0124^{**}$ & -127 \\
\hline
saclay      &  $0.0089^{*}$   &  740  & conseil       & $-0.0116^{**}$ & -626 \\
\hline
quartier    &  $0.0086^{*}$   &   40  & militant      & $-0.0112^{**}$ & -47  \\
\hline
\end{tabular}}
\caption{Top 10 absolute values of $\Delta \mathrm{TFIDF}(w)$ for \tp. Words with positive values are more characteristic of converted outliers (\(\mathcal{H}\)), and those with negative values are more typical of non-converted outliers (\(\text{not } \mathcal{H}\)). Statistical significance is based on the Kruskal--Wallis test; $^{*}$ and $^{**}$ indicate $p$-values $< 0.05$ and $< 0.01$, respectively. $\Delta \mathrm{Occ}(w)$ indicates the difference in word occurrence counts between the two groups.}
\label{tab:tfidf-tp}
\end{table*}

\subsection{Replication Study Appendix}
\label{sec:replicappendix}

\subsubsection{Validation or invalidation of \h per model over different periods for GHG}
\label{sec:timereplicappendix}

For a detailed examination of model variations, Table \ref{tab:timeconversionghg} presents the validation or invalidation of \h for each model over different periods of cumulative clustering. Among English models, ${\small\texttt{distilbert-base-uncased}}$, ${\small\texttt{e5-base-v2}}$, and ${\small\texttt{all-MiniLM-L12-v2}}$ show complete consistency, achieving full integration early. ${\small\texttt{all-roberta-large-v1}}$ follows a steady trajectory, with conversion decreasing slightly from 93.62\% to 88.17\%. Among French models, ${\small\texttt{sentence-camembert-base}}$, a French model, shows instability, with a conversion fluctuating from 46.43\% to 58.14\% before dropping to 43.18\%. ${\small\texttt{Solon-embeddings-large-0.1}}$, despite being a French model, integrates all outliers early, aligning with its high absolute validation score. Multilingual models exhibit mixed behaviors., with ${\small\texttt{multilingual-e5-large}}$ achieving full integration like English models, while ${\small\texttt{paraphrase-multilingual-mpnet-base-v2}}$ and ${\small\texttt{xlm-roberta-large}}$ retain substantial outliers (with 40.70\% and 43.91\%, respectively). ${\small\texttt{multilingual-mpnet-base-v2}}$ initially increases its conversion (26.32\% to 38.10\%) before stabilizing. ${\small\texttt{xlm-roberta-large}}$ exhibits a downward trend, with conversion dropping from 45.00\% to 22.00\%.  

\begin{table*}[t]
\centering
\resizebox{\textwidth}{!}{
\begin{tabular}{llcccc}
\toprule
\multirow{2}{*}{\textbf{Model}} & \multirow{2}{*}{\textbf{Measures}} & \multicolumn{4}{c}{\textbf{Time}} \\
\cmidrule{3-6}
& & \textbf{2022-10 (50\%)}& \textbf{2023-02 (70\%)}& \textbf{2023-06 (90\%)}& \textbf{Remaining (100\%)}\\
\midrule
\multirow{3}{*}{${\texttt{Solon-embeddings-large-0.1}}$} 
& Nb Outliers / All Articles at $t$ & 79/79& 18/105& 96/236& 0/312\\
& \% Becoming Inliers at $(t+n)$ & 100\% & 100\% & 100\% & Converted on 2023-07 \\
\midrule
\multirow{3}{*}{${\texttt{...-multi..-mpnet-base-v2}}$} 
& Nb Outliers / All Articles at $t$ & 19/79& 21/105& 81/236& 127/312\\
& \% Becoming Inliers at $(t+n)$ & 26.32\% & 38.10\% & 33.33\% & - \\
\midrule
\multirow{3}{*}{${\texttt{sentence-camembert-base}}$} 
& Nb Outliers / All Articles at $t$ & 28/79& 43/105& 88/236& 80/312\\
& \% Becoming Inliers at $(t+n)$ & 46.43\% & 58.14\% & 43.18\% & - \\
\midrule
\multirow{3}{*}{${\texttt{multi..-e5-large}}$} 
& Nb Outliers / All Articles at $t$ & 23/79& 26/105& 49/236& 0/312\\
& \% Becoming Inliers at $(t+n)$ & 100\% & 100\% & 100\% & Converted on 2023-07 \\
\midrule
\multirow{3}{*}{${\texttt{xlm-roberta-large}}$} 
& Nb Outliers / All Articles at $t$ & 20/79& 60/105& 76/236& 137/312\\
& \% Becoming Inliers at $(t+n)$ & 45.00\% & 45.00\% & 22.00\% & - \\
\midrule
\multirow{3}{*}{${\texttt{all-MiniLM-L12-v2}}$} 
& Nb Outliers / All Articles at $t$ & 79/79& 69/105& 90/236& 0/312\\
& \% Becoming Inliers at $(t+n)$ & 100\% & 100\% & 100\% & Converted on 2023-07 \\
\midrule
\multirow{3}{*}{${\texttt{all-roberta-large-v1}}$} 
& Nb Outliers / All Articles at $t$ & 47/79& 40/105& 93/236& 32/312\\
& \% Becoming Inliers at $(t+n)$ & 93.62\% & 92.50\% & 88.17\% & - \\
\midrule
\multirow{3}{*}{${\texttt{distilbert-base-uncased}}$} 
& Nb Outliers / All Articles at $t$ & 42/79& 33/105& 87/236& 0/312\\
& \% Becoming Inliers at $(t+n)$ & 100\% & 100\% & 100\% & Converted on 2023-07 \\
\midrule
\multirow{3}{*}{${\texttt{e5-base-v2}}$} 
& Nb Outliers / All Articles at $t$ & 13/79& 27/105& 58/236& 0/312\\
& \% Becoming Inliers at $(t+n)$ & 100\% & 100\% & 100\% & Converted on 2023-07 \\
\bottomrule
\end{tabular}}
\caption{Proportion of outliers converting to clusters in \ghg, for each model and along four time windows.}
\label{tab:timeconversionghg}
\end{table*}

That said, trends across models reveal a broadly consistent trajectory: high early conversion of outliers into topic inliers (ranging from 26.32\% to 100\% in 2022-10), followed by a mid-phase slowdown with moderate-to-low integration (10.95\%–65.71\% in 2023-02), and eventual stabilization with persistent outliers in the final stage (10.25\%–43.91\%). Most models adhere to this pattern, with strong early conversion seen in ${\small\texttt{all-roberta-large-v1}}$ (93.62\%) and ${\small\texttt{e5-base-v2}}$ (100\%), followed by a gradual decline in mid-phase integration for models like ${\small\texttt{sentence-camembert-base}}$ (fluctuating from 46.43\% to 58.14\%) and ${\small\texttt{multilingual-mpnet-base-v2}}$ (increasing from 26.32\% to 38.10\%). By the final stage, outlier retention converges to similar rates across models, such as ${\small\texttt{sentence-camembert-base}}$ stabilizing at 25.64\% and ${\small\texttt{xlm-roberta-large}}$ retaining 43.91\% of outliers.

\subsubsection{Top 10 Distinguishing Terms Based on TF-IDF Differences Between Outliers Validating and Not Validating \h}
\label{sec:tfidfreplicappendix}

Table~\ref{tab:tfidf-ghg} lists the top 10 terms whose TF-IDF scores most strongly differentiate outliers that validate hypothesis \h from those that do not, highlighting key lexical features associated with each group.

\renewcommand{\arraystretch}{1.15}
\begin{table*}[t]
\centering
\resizebox{0.7\textwidth}{!}{%
\begin{tabular}{|l|r|r||l|r|r|}
\hline
\textbf{Word} &  $\boldsymbol{\Delta \mathrm{TFIDF}(w)}$ & $\boldsymbol{\Delta \mathrm{Occ}(w)}$ &
\textbf{Word} &  $\boldsymbol{\Delta \mathrm{TFIDF}(w)}$ & $\boldsymbol{\Delta \mathrm{Occ}(w)}$ \\
\hline
climate     &  0.0067          &  17851  & amazon     & -0.0034         & -98   \\
\hline
report      &  0.0051$^{*}$    &  2656   & pakistan   & -0.0033         & -130  \\
\hline
degree      &  0.0035          &  2576   & china      & -0.0031         & -480  \\
\hline
said        &  0.0035          &  9134   & child      & -0.0027         & -227  \\
\hline
bill        &  0.0033$^{*}$    &  804    & thunberg   & -0.0024         & -63   \\
\hline
company     &  0.0032          &  2577   & reactor    & -0.0023         & -65   \\
\hline
would       &  0.0031          &  3668   & protest    & -0.0023         & -87   \\
\hline
republican  &  0.0030          &  728    & soil       & -0.0023         & -185  \\
\hline
energy      &  0.0030          &  5651   & granholm   & -0.0023         & -51   \\
\hline
nice        &  0.0029          &  895    & art        & -0.0023         & -172  \\
\hline
\end{tabular}}
\caption{Top 10 absolute values of \(\Delta \text{TFIDF}(w)\) for \ghg. Words with positive values are more characteristic of converted outliers (\(\mathcal{H}\)); words with negative values are more typical of non-converted outliers (\(\text{not } \mathcal{H}\)). Statistical significance is based on the Kruskal-Wallis test; $^{*}$ indicates $p < 0.05$. \(\Delta \text{Occ}(w)\) shows the difference in word frequency between the two groups.}
\label{tab:tfidf-ghg}
\end{table*}

\end{document}